\documentclass[10pt,twocolumn,letterpaper]{article}

\usepackage[accsupp]{axessibility}
\usepackage{cvpr}              %

\usepackage[table]{xcolor}

\usepackage{multirow}

\usepackage{colortbl}

\usepackage{booktabs}

\usepackage{bm}

\usepackage{graphicx}

\usepackage{caption}

\usepackage{array}

\usepackage{amssymb}

\usepackage{amsmath}

\usepackage{wrapfig}

\usepackage{booktabs,makecell,multirow,pifont}

\usepackage{xspace}

\definecolor{cvprblue}{rgb}{0.21,0.49,0.74}
\usepackage[pagebackref,breaklinks,colorlinks,allcolors=cvprblue]{hyperref}

\usepackage{marvosym}
\newcommand{\corrmark}{\begingroup\renewcommand\thefootnote{\Letter}\footnotemark\endgroup}

\newcommand{\modelname}{LegoOcc}

\def\paperID{} %
\def\confName{CVPR}
\def\confYear{2026}

\title{Monocular Open Vocabulary Occupancy Prediction for Indoor Scenes}

\author{
Changqing Zhou$^{1}$ \quad
Yueru Luo$^{2}$ \quad
Han Zhang$^{1}$ \quad
Zeyu Jiang$^{1}$ \quad
Changhao Chen$^{1}$\corrmark
\vspace{3pt} \\
$^{1}$ The Hong Kong University of Science and Technology (Guangzhou) \\
$^{2}$ The Chinese University of Hong Kong, Shenzhen \\
{\tt\small czhou149@connect.hkust-gz.edu.cn \quad changhaochen@hkust-gz.edu.cn}
}

\begin{document}
\maketitle
\begingroup\renewcommand\thefootnote{\Letter}\footnotetext{Corresponding author.}\endgroup

\definecolor{ceiling}{RGB}{214,  38, 40}   %
\definecolor{floor}{RGB}{43, 160, 4}     %
\definecolor{wall}{RGB}{158, 216, 229}  %
\definecolor{window}{RGB}{114, 158, 206}  %
\definecolor{chair}{RGB}{204, 204, 91}   %
\definecolor{bed}{RGB}{255, 186, 119}  %
\definecolor{sofa}{RGB}{147, 102, 188}  %
\definecolor{table}{RGB}{30, 119, 181}   %
\definecolor{tvs}{RGB}{160, 188, 33}   %
\definecolor{furniture}{RGB}{255, 127, 12}  %
\definecolor{objects}{RGB}{196, 175, 214} %

\definecolor{Sbg}{HTML}{F4F9FF}        %
\definecolor{Ubg}{HTML}{FFF7F0}        %
\definecolor{SbgStrong}{HTML}{DCEBFF}  %
\definecolor{UbgStrong}{HTML}{FFE2CC}  %
\definecolor{SbgVivid}{HTML}{91b4ed} 
\definecolor{UbgVivid}{HTML}{f2aa61}

\newcommand{\cmark}{\ding{51}}

\begin{abstract}
Open-vocabulary 3D occupancy is vital for embodied agents, which need to understand complex indoor environments where semantic categories are abundant and evolve beyond fixed taxonomies. 
While recent work has explored open-vocabulary occupancy in outdoor driving scenarios, such methods transfer poorly indoors, where geometry is denser, layouts are more intricate, and semantics are far more fine-grained.
To address these challenges, we adopt a geometry-only supervision paradigm that uses only binary occupancy labels (occupied vs.\ free).
Our framework builds upon 3D Language-Embedded Gaussians, which serve as a unified intermediate representation coupling fine-grained 3D geometry with a language-aligned semantic embedding.
On the geometry side, we find that existing Gaussian-to-Occupancy operators fail to converge under such weak supervision, and we introduce an opacity-aware, Poisson-based approach that stabilizes volumetric aggregation.
On the semantic side, direct alignment between rendered features and open-vocabulary segmentation features suffers from feature mixing; we therefore propose a Progressive Temperature Decay schedule that gradually sharpens opacities during splatting, strengthening Gaussian-language alignment. On Occ-ScanNet, our framework achieves 59.50 IoU and 21.05 mIoU in the open-vocabulary setting, surpassing all existing occupancy methods in IoU and outperforming prior open-vocabulary approaches by a large margin in mIoU. Code will be released at \url{https://github.com/JuIvyy/LegoOcc}.
\end{abstract}

\section{Introduction}\label{sec:intro}

Comprehensive 3D geometric and semantic understanding of surrounding environments is fundamental for embodied agents across diverse applications such as service robots, drones, and AR/VR systems. Within this context, \emph{semantic occupancy} has emerged as a powerful and compact formulation that unifies dense geometry and semantics within a volumetric grid, enabling high-level spatial reasoning and downstream embodied decision-making~\cite{volumetrivln,occvla,occllama}.
Vision-centric occupancy prediction is particularly attractive due to the ubiquity, low cost, and rich appearance cues of cameras. Recent studies and benchmarks show that vision-only pipelines can infer competitive 3D occupancy from both \emph{monocular} inputs~\cite{ISO,embodiedocc,embodiedocc++,yu2025shtocc,OccFormer,VoxFormer,SparseOcc} and \emph{multi-view} inputs~\cite{surroundocc,gaussianformer,gaussianformer2,gaussrender}, employing architectures ranging from transformer-based to Gaussian-splatting-based designs.%

\begin{figure}[t]
  \centering
  \includegraphics[width=0.5\textwidth]{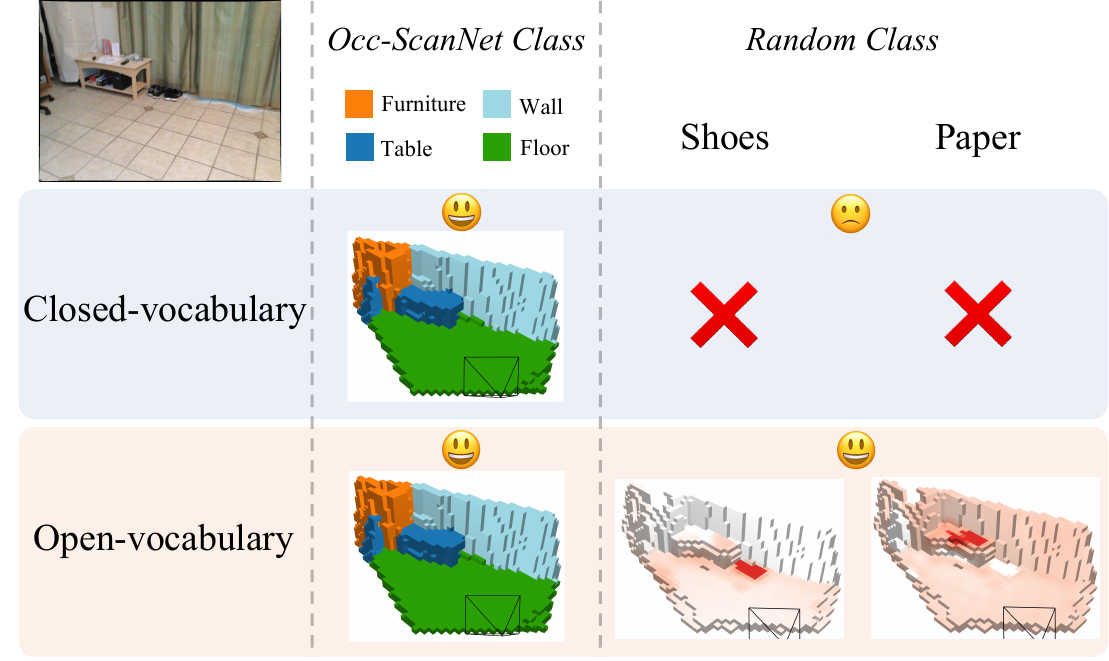}
  \caption{\textbf{Closed- vs.\ open-vocabulary occupancy.} Prior methods~\cite{ISO,embodiedocc} trained under a \emph{closed vocabulary} can label only the categories predefined at training time, which restricts real-world deployment. Our \emph{open-vocabulary} approach aligns language with 3D occupancy and supports text queries for arbitrary categories. \emph{Right column (Random Class):} text-conditioned per-voxel scores are visualized as heatmaps; darker red indicates higher likelihood for the queried category.}
  \label{fig:teaser}
  \vspace{-5mm}
\end{figure}

In general, occupancy prediction involves two key components: \textit{geometry learning}, which determines whether a voxel is occupied, and \textit{semantic learning}, which assigns a category to each occupied voxel. However, real-world deployments often encounter highly long-tailed category distributions~\cite{zhang2023deeplongtailsurvey} and inherently open-ended semantic spaces~\cite{joseph2021towards}. Consequently, restricting predictions to a fixed, closed vocabulary limits practical utility and motivates the development of open-vocabulary occupancy prediction. As shown in~\cref{fig:teaser}, conventional closed-vocabulary models can only recognize training-time categories, whereas open-vocabulary formulations lift this restriction and significantly broaden real-world applicability.

This direction has progressed rapidly in outdoor driving scenarios, where recent methods integrate vision–language models (VLMs) and text prompts to associate camera-based 3D occupancy with free-form semantics~\cite{veon,yu2025language,boeder2024langocc,li2025ago,groundingocc}. By contrast, indoor occupancy prediction remains relatively underexplored and is typically trained or evaluated under small, closed taxonomies; for instance, Occ-ScanNet~\cite{ISO} defines only 11 semantic classes, which underrepresents the rich and diverse semantics of indoor environments. 
Indoor scenes further differ fundamentally from outdoor ones in \textit{two critical aspects}: \textbf{1)} they exhibit much denser and more complex geometry, and \textbf{2)} they contain far more fine-grained, long-tailed semantic categories. As a result, directly applying outdoor open-vocabulary occupancy pipelines to indoor settings yields limited performance~\cite{embodiedocc}, as also evidenced in our experiments (see ~\cref{tab:main_mono} and discussion in~\cref{sec:main_results}).
Early attempts such as OVO~\cite{ovo} take an initial step by distilling 2D open-vocabulary segmentation into 3D voxels on smaller-scale datasets, but still rely on a fixed set of base categories during training. These limitations highlight both the practical importance and the open challenge of indoor open-vocabulary occupancy: embodied agents must reason about arbitrary objects and spatial layouts beyond a fixed label set before taking action, yet existing indoor methods fall short of this requirement.

To address this gap, we adopt a geometry-only supervision protocol: training uses only binary voxel occupancy labels and no semantic voxel annotations, and we tackle the resulting challenges on both the geometric and semantic fronts. This setting is practically motivated, as geometry can be harvested at scale with modest effort. Specifically, ScanNet provides images and depth captured by an RGB-D camera, and Occ-ScanNet generates occupancy ground truth via the SCFusion pipeline~\cite{scfusion}, which incrementally fuses occupancy probabilities from depth-based reconstruction~\cite{supereight} and completion methods~\cite{scan2cad,forknet}. In particular, SCFusion fuses depth with aligned ShapeNet CAD models via Scan2CAD~\cite{scan2cad} to recover complete-scene geometry and mitigate occlusions inherent in raw sensor data. In contrast, semantic labels remain expensive to curate due to the high diversity and long-tailed distribution of indoor categories.

To enable open-vocabulary occupancy prediction under geometry-only supervision, we employ 3D Language-Embedded Gaussians (LE-Gaussians)~\cite{shi2023legaussians,zhou2024featuregs} as the spatial intermediate representation. Each LE-Gaussian couples (1) native geometric parameters—position, rotation, scale, and opacity—that encode local occupancy evidence, with (2) a learnable semantic embedding that anchors language-aligned~\cite{clip} semantics at a specific 3D location.
\textbf{On the geometry side}, indoor environments contain dense structures and severe occlusions, demanding fine-grained volumetric modeling. While Gaussian primitives offer an expressive and computationally efficient representation, existing Gaussian-to-Occupancy (G2O) mappings become unstable when trained solely with binary occupancy labels due to insufficient opacity modeling (\cref{sec:g2o}).
To address this, we introduce an opacity-aware, \emph{Poisson-based Gaussian-to-Occupancy} formulation. Our model treats each Gaussian’s effective opacity as a nonnegative event intensity and interprets voxel occupancy as the first-arrival event of a Poisson process, enabling stable and principled volumetric aggregation.
\textbf{On the semantic side}, indoor categories are highly fine-grained and long-tailed, causing naïve alignment between rendered Gaussian features and 2D open-vocabulary segmentations to suffer from feature mixing when multiple categories overlap in image space (\cref{sec:gs_temp}). Concretely, we render LE-Gaussian features into images via Gaussian splatting~\cite{gaussiansplatting} and align them with features extracted by a training-free open-vocabulary segmenter (e.g., Trident~\cite{trident}).
To mitigate feature dilution caused by weighted feature aggregation, we introduce a \emph{Progressive Temperature Decay} schedule that gradually sharpens effective opacities during splatting. This suppresses cross-category blending and yields more discriminative, language-aligned 3D features for indoor open-vocabulary occupancy prediction.

In summary, our contributions are threefold:
\begin{itemize}
\item We introduce \textbf{\modelname}, which leverages \textbf{L}anguage-\textbf{e}mbedded \textbf{G}aussians as a fine-grained 3D spatial intermediate representation for monocular \textbf{O}pen-vocabulary \textbf{Occ}upancy prediction in large-scale indoor environments, enabling embodied agents to reason about arbitrary objects beyond closed taxonomies.
\item We propose an opacity-aware, Poisson-based Gaussian-to-Occupancy operator that operates reliably under geometry-only (binary) occupancy supervision. In addition, we develop a Progressive Temperature Decay schedule that sharpens Gaussian opacities during splatting, thereby enhancing Gaussian–language feature alignment.
\item We conduct extensive experiments on Occ-ScanNet, where our framework achieves 59.50 IoU and 21.05 mIoU in the open-vocabulary setting, surpassing the previous SoTA by 3.02 IoU and outperforming prior open-vocabulary methods by 11.80 mIoU (more than $2\times$ the best previous result); ablation studies further validate the effectiveness of each proposed component.
\end{itemize}

\section{Related Works}\label{sec:related_work}

\subsection{Occupancy Prediction}

Camera-only occupancy has been extensively studied in outdoor autonomous driving, often using grid- or transformer-based pipelines~\cite{gaussianformer,gaussianformer2,occupancypoints}, whereas indoor settings remain comparatively underexplored. MonoScene first cast semantic scene completion 
as an occupancy prediction task~\cite{monoscene,sscnet,ndcscene}, and subsequent works on indoor datasets such as Occ-ScanNet~\cite{ISO} improved
accuracy and scalability. Notable methods include ISO~\cite{ISO} with depth-guided feature lifting, EmbodiedOcc~\cite{embodiedocc} with Gaussian-based volumetric prediction, and RoboOcc~\cite{roboocc} with opacity-guided encoders.
Methodologically, indoor approaches span dense volumetric lifting of 2D features along rays into voxel grids followed by decoding with 3D CNNs or volumetric heads~\cite{monoscene,ndcscene,ISO,fbocc,lss}.
More recent works employ
transformer-augmented volumetrics and sparse Gaussian/point primitives that iteratively refine geometry–semantics coupling for efficiency and fidelity~\cite{gaussianformer,embodiedocc,embodiedocc++,roboocc}.
However, indoor environments differ markedly from structured road scenes in layout complexity, scale/viewpoint variation, and long-tailed distribution of fine-grained categories.
Consequently, designs tailored for driving scenes often transfer suboptimally indoors, motivating a re-examination of feature lifting, spatial representation, and supervision paradigms~\cite{ISO,embodiedocc,roboocc}.

\subsection{Open Vocabulary Occupancy Prediction}

Open-vocabulary occupancy has recently gained traction in autonomous driving as a way to mitigate the cost and difficulty of annotating 3D occupancy. These methods typically adopt self-/weakly supervised settings to learn voxel-level geometry and semantics without full 3D labels.
For semantic learning, most approaches first derive open-vocabulary cues from VLMs and obtain pseudo 2D masks with open-set segmentors~\cite{groundedsam,groundingdino,jiang2026freeocc}. These signals are then either lifted to 3D by projecting pixels onto point clouds or associating pixels/points with voxels~\cite{occnerf,li2025ago,yu2025language}, or mapped to voxels via differentiable voxel projection~\cite{veon,pop3d}, or used to supervise a rendered 3D semantic field directly in the image plane via volume rendering~\cite{boeder2024langocc}. 
Feature alignment is commonly achieved through cosine similarity between voxel features and text/image embeddings~\cite{groundedsam,groundingdino}.
For geometry learning, two families dominate: LiDAR-derived pseudo labels aggregated into voxels for supervision~\cite{veon,gao2025loc,pop3d,li2025ago}, and self-supervised multi-view rendering to learn occupancy by enforcing photometric/feature consistency across frames without 3D ground truth~\cite{selfocc,occnerf}.

In indoor scenarios, research remains limited. OVO~\cite{ovo} makes an early attempt on a small-scale dataset by distilling open-vocabulary semantics into a 3D occupancy network via pixel-wise alignment between projected voxel features and 2D open-vocabulary segmentation.
OpenOcc~\cite{OpenOcc} further
leverages neural radiance fields to unifying the 3D scene reconstruction and understanding from RGB-D inputs, while
OG~\cite{dong2023og} integrates grounded-SAM~\cite{groundedsam} with affinity fields to distinguish grounded instances.
Despite these efforts, open-vocabulary occupancy for large-scale indoor scenes remains largely underexplored.

\section{Method}\label{sec:method}

\subsection{Problem Setting}\label{sec:overview}
\begin{figure*}[h]
  \centering
  \includegraphics[width=0.98\textwidth]{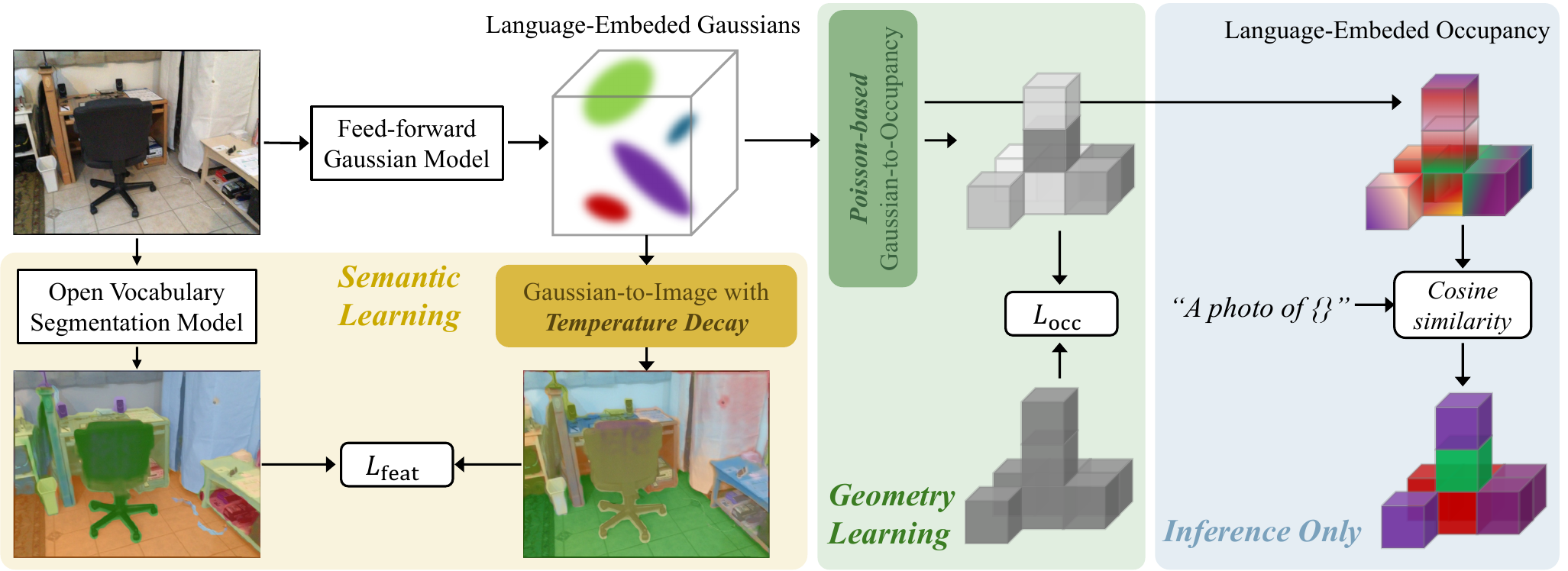}
  \vspace{-1mm}
  \caption{\textbf{\modelname~Framework Overview.} From a monocular image, a feed-forward Gaussian model produces \emph{Language-Embedded Gaussians}. Training proceeds along two paths: \emph{Semantic learning}, we differentiably render Gaussian features to the image with \emph{Progressive Temperature Decay} and align them to a training-free open-vocabulary segmenter via a cosine objective $L_{\text{feat}}$; \emph{Geometry learning}, we convert Gaussians to occupancy using an opacity-aware Poisson-based Gaussian-to-Occupancy operator and supervise binary occupancy with $L_{\text{occ}}$. At inference, the Language-Embedded Occupancy supports text-driven queries by computing cosine similarity between voxel embeddings and prompt embeddings, yielding open-vocabulary semantic occupancy without dense voxel-level semantic labels during training.}
  \label{fig:framework}
  \vspace{-3mm}
\end{figure*}

Given an input image $\mathbf{I}\in\mathbb{R}^{H\times W\times 3}$, the goal of occupancy estimation is to predict a semantic occupancy field $\mathbf{O}\in\mathbb{R}^{X\times Y\times Z\times C}$, where $(X,Y,Z)$ specify the resolution of the local 3D volume and $C$ is the number of semantic categories. We decompose the problem into two complementary components: (i) \emph{geometry learning}, which determines whether each voxel is occupied, and (ii) \emph{semantic learning}, which assigns a category to occupied voxels.

To enable open-vocabulary occupancy, we employ \emph{Language-Embedded Gaussians} (LE-Gaussians) as a unified 3D intermediate representation coupling geometry and language-aligned semantics. 
Each LE-Gaussian is parameterized as:
\begin{equation}\label{eq:gauss_3d}
\mathcal{G}_i=\big(\boldsymbol{\mu}_i,\boldsymbol{\Sigma}_i,\alpha_i,\mathbf{f}_i\big),
\end{equation}
where $\boldsymbol{\mu}_i\in\mathbb{R}^3$ and $\boldsymbol{\Sigma}_i\in\mathbb{R}^{3\times 3}$ are the Gaussian center and covariance, $\alpha_i\in[0,1]$ denotes opacity, and $\mathbf{f}_i\in\mathbb{R}^d$ is a language-aligned embedding with $d$ channels.
LE-Gaussians thus explicitly bind \emph{geometry} (via $\boldsymbol{\mu}_i,\boldsymbol{\Sigma}_i,\alpha_i$) and \emph{open-vocabulary semantics} (via $\mathbf{f}_i$) within a single set of primitives, supporting both occupancy prediction and semantic reasoning.

\subsection{\modelname~Framework Overview}\label{sec:overview}
As illustrated in~\cref{fig:framework}, given an input image $\mathbf{I}$, a feed-forward Gaussian predictor outputs a set of LE-Gaussians $\mathbf{G}=\{\mathcal{G}_i\}_{i=1}^{N}$. The same set $\mathbf{G}$ is then reused for two purposes: 1) \emph{occupancy prediction} (\cref{sec:g2o}), where our Poisson-based Gaussian-to-Occupancy (G2O) operator infers 3D occupancy from geometric parameters $(\boldsymbol{\mu}_i,\boldsymbol{\Sigma}_i,\alpha_i)$; and
2) \emph{language-aligned semantic prediction}, where per-Gaussian embeddings $\mathbf{f}_i$ are used to infer the semantic label of each occupied voxel. Supervision is imposed by rendering these embeddings to the image plane and aligning the rendered features with those from open-vocabulary segmentation models; the alignment is further improved by our progressive temperature decay (\cref{sec:gs_temp}).
Training jointly minimizes a supervised binary-occupancy loss with ground-truth and a rendered-feature alignment loss that requires no 2D human annotations; see \cref{sec:loss} for details.

\subsection{Poisson-based Gaussian-to-Occupancy}\label{sec:g2o}

3D Gaussians offer an efficient and expressive representation to capture fine-grained indoor geometry
and have been widely adopted for scene reconstruction. In this section, we first revisit Gaussian splatting, analyze why existing G2O formulations become problematic under binary-only occupancy supervision, and then introduce our Poisson-based G2O reformulation that addresses the issue.

\noindent\textbf{Gaussian Splatting.}
We follow the front-to-back $\alpha$-blending strategy of Gaussian Splatting~\cite{gaussiansplatting} to render features from LE-Gaussians.
For the $i$-th Gaussian $\mathcal{G}_i$ with feature $\mathbf{f}_i$, opacity $\alpha_i$, and screen-space mean/covariance $(\boldsymbol{\mu}'_i,\boldsymbol{\Sigma}'_i)$, its 2D kernel at image location $\mathbf{x}'$ is
\begin{equation}
p_i(\mathbf{x}') \;=\; \exp\!\Big(-\tfrac{1}{2}\,(\mathbf{x}'-\boldsymbol{\mu}'_i)^{\!\top}\,{\boldsymbol{\Sigma}'_i}^{-1}\,(\mathbf{x}'-\boldsymbol{\mu}'_i)\Big),
\end{equation}
where primes denote screen-space quantities corresponding to the 3D parameters in \cref{eq:gauss_3d}.
We define the \emph{effective opacity} at $\mathbf{x}'$ as
\begin{equation}
\tilde{\alpha}_i(\mathbf{x}') \;=\; \alpha_i\, p_i(\mathbf{x}').
\end{equation}
and render the pixel feature by $\alpha$-blending:
\begin{equation}
\mathbf{F}(\mathbf{x}')
\;=\;
\sum_{i=1}^{N}
\Bigg(\prod_{j=1}^{i-1}\!\big(1-\tilde{\alpha}_j(\mathbf{x}')\big)\Bigg)\,
\tilde{\alpha}_i(\mathbf{x}')\,\mathbf{f}_i.
\label{eq:a_blend}
\end{equation}

\noindent\textbf{Limitations of GaussianFormer2.}
To infer occupancy from Gaussians, GaussianFormer2~\cite{gaussianformer2} introduces a \emph{probabilistic Gaussian superposition} that treats each primitive $\mathcal{G}_i$ as a local occupancy distribution and aggregates them multiplicatively. Concretely, the contribution of $\mathcal{G}_i$ at a 3D location $\mathbf{x}$ is
\begin{equation}
p_i(\mathbf{x}) \;=\; \exp\!\Big(-\tfrac{1}{2}\,(\mathbf{x}-\boldsymbol{\mu}_i)^{\!\top}\,\boldsymbol{\Sigma}_i^{-1}\,(\mathbf{x}-\boldsymbol{\mu}_i)\Big),
\label{eq:single-prob}
\end{equation}
and, assuming independence across Gaussians, the aggregated occupancy becomes
\begin{equation}
p(\mathbf{x}) \;=\; 1 - \prod_{i=1}^{N}\!\big(1 - p_i(\mathbf{x})\big).
\label{eq:multi-prob}
\end{equation}
While efficient and simple, this formulation omits opacity $\alpha_i$ in the geometry branch and relies solely on the spatial kernels.
As a result, the occupancy inference is decoupled from the opacity used in image rendering, creating a mismatch between 2D rendering and 3D voxel aggregation.

\noindent\textbf{Why is this problematic?}
In GaussianFormer2, geometry is aggregated without opacity (Eq.~\ref{eq:multi-prob}), but this does not cause issues because: 1) their semantic learning explicitly models opacity, and 2) it does not rely on 2D rendering during training, so no inconsistency arises.
By contrast, our semantics are learned by \emph{rendering} LE-Gaussian (Eq.~\ref{eq:a_blend}) to images and aligning them to 2D open-vocabulary segmentations features, which is inherently opacity-aware.
Using an opacity-agnostic voxel aggregation for geometry while supervising semantics via opacity-aware rendering introduces conflicting signals and unstable optimization.

\noindent\textbf{A Bernoulli approach.}
A direct fix is to replace the spatial kernel $p_i(\mathbf{x})$ with the \emph{effective opacity} $\tilde{\alpha}_i(\mathbf{x})=\alpha_i\,p_i(\mathbf{x})$ and reuse the complementary-probability rule:
\begin{equation}
p(\mathbf{x})
= 1-\prod_{i=1}^{N}\!\big(1-\tilde{\alpha}_i(\mathbf{x})\big)
= 1-\prod_{i=1}^{N}\!\big(1-\alpha_i\,p_i(\mathbf{x})\big),
\label{eq:naive_opacity}
\end{equation}
which can be viewed as independent Bernoulli “hits’’ from overlapping Gaussians at $\mathbf{x}$.
A similar opacity-based idea appears in Gaussian Opacity Fields~\cite{Yu2024GOF} and in concurrent fully supervised closed-vocabulary method~\cite{qian2025splatsscdecoupleddepthguidedgaussian,park2025s2gostreamingsparsegaussian}.

However, we observe that the Bernoulli aggregation drives the learned opacities to \emph{small} values, which enlarges the gap between rendered features and per-Gaussian embeddings (see \cref{sec:gs_temp}).
This behavior likely arises because when multiple Gaussians overlap within a voxel, the Bernoulli union $1-\prod_i(1-\alpha_i p_i)$ saturates rapidly toward~1; and for a fixed total mass $\sum_i \alpha_i p_i$, the union is maximized when each of the overlapping contributions are similar, making it difficult to learn differentiated opacities.

\paragraph{A Poisson approach.}
We adopt a Poisson formulation that treats each Gaussian’s local contribution as a nonnegative \emph{event intensity} (mean count):
\begin{equation}
h_i(\mathbf{x}) \;\triangleq\; \alpha_i\,p_i(\mathbf{x}) \;\ge 0,
\qquad
z(\mathbf{x}) \;=\; \sum_{i=1}^{N} h_i(\mathbf{x}) ,
\label{eq:hgs_hz}
\end{equation}
where $h_i$ is the expected contribution of Gaussian $\mathcal{G}_i$ over a small voxel segment, and $z(\mathbf{x})$ is the \emph{mean measure} (cumulative expected count) from all Gaussians at $\mathbf{x}$.
Modeling occupancy as the probability that a non-homogeneous Poisson process (NHPP) has produced at least one event at location $\mathbf{x}$ gives a simple closed form.
Let $z(\mathbf{x})$ denote the accumulated mean measure at $\mathbf{x}$.
A classical result for Poisson processes states that the probability of observing no event is
$\Pr[\text{no event at }\mathbf{x}] = \exp\!\big(-z(\mathbf{x})\big)$~\cite{ross1996stochastic,kingman-poisson-processes}.
Interpreting “at least one event” as the space being occupied, the occupancy becomes a single exponential of the total mean measure:
\begin{equation}
p(\mathbf{x})
= 1-\exp\!\big(-z(\mathbf{x})\big)
= 1-\exp\!\Big(-\sum_{i=1}^{N}\alpha_i\,p_i(\mathbf{x})\Big).
\label{eq:hgs_occ}
\end{equation}

\subsection{Progressive Temperature Decay}\label{sec:gs_temp}

As shown in the semantic learning part of~\cref{fig:framework}, we adopt Gaussian Splatting~\cite{gaussiansplatting} to render LE-Gaussians into images. 
However, the $\alpha$-blending in~\cref{eq:a_blend} produces a \emph{weighted sum} of per-ray contributions, so each pixel feature becomes a mixture of multiple Gaussian embeddings along the ray. Consequently, the supervision signal encourages the rendered mixture, rather than each individual Gaussian, to be language-aligned, creating a mismatch that later hampers G2O-based splatting. This ambiguity is further exacerbated in indoor scenes, where objects are numerous, fine-grained, and heavily overlapping in projection.
Dr.\ Splat~\cite{drsplat25} mitigates this via hard \emph{feature registration}: along a pixel ray it assigns the pixel’s CLIP embedding to only the Top-$k$ dominant Gaussians, yielding language-aligned vectors that are directly retrievable in 3D. While effective, this discrete selection causes sparse gradients, since supervision backpropagates only through the chosen Top-$k$ primitives.

\begin{figure}[h]
  \centering
  \includegraphics[width=0.4\textwidth]{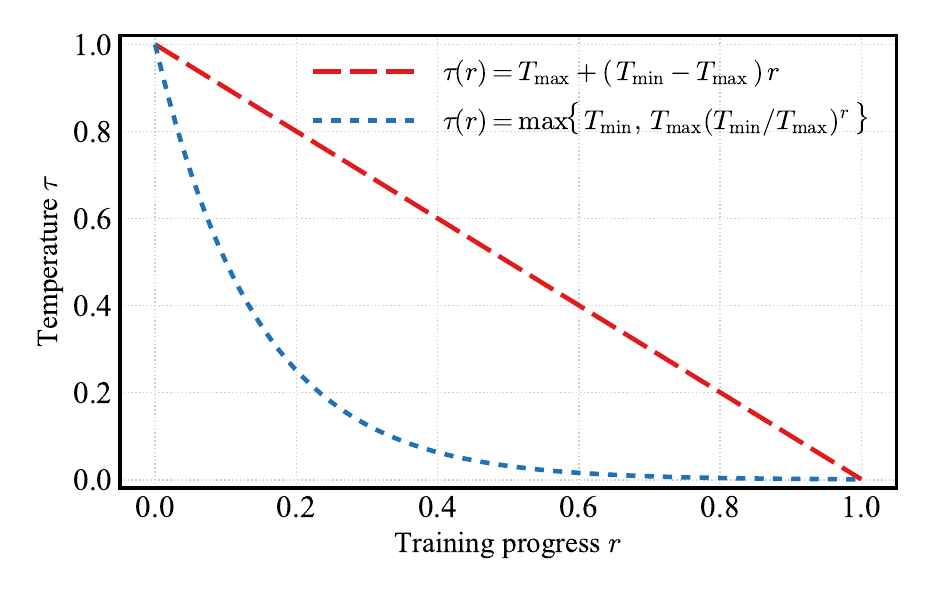}
  \vspace{-2mm}
 \caption{\textbf{Comparison of temperature schedules.} Linear decay decreases $\tau$ uniformly, whereas our exponential schedule rapidly approaches $T_{\min}$, allocating more iterations for the model to adapt to the low-temperature regime.}
  \vspace{-3mm}
  \label{fig:temp_decay}
\end{figure}

Instead of a hard Top-$k$, we reduce mixture ambiguity via a continuous and training-stable \emph{opacity sharpening}.
Let $\alpha_i=\sigma(\alpha_i^{\text{logit}})$, where $\sigma$ is the sigmoid function and $\alpha_i^{\text{logit}}$ is the predicted logit. We adopt a \emph{tempered} sigmoid
\begin{equation}
\alpha_i \;=\; \sigma\!\left(\frac{\alpha_i^{\text{logit}}}{\tau}\right),
\end{equation}
where smaller (near 0) temperature $\tau$ values sharpen opacities
towards $\{0,1\}$, reducing feature mixing along the ray while maintaining gradient flow to all contributors (rather than a discrete subset).
In practice, we employ a \emph{progressive temperature decay} schedule during training, annealing $\tau$ from $T_{\max}$ to $T_{\min}$ (default $T_{\max}{=}1$, $T_{\min}{=}10^{-3}$). Let $r\!\in[0,1]$ denote the normalized training progress, 0 means beginning, 1 means end. We use an \emph{exponential} schedule
\[
\tau(r)\;=\;\max\!\Big\{T_{\min},\;T_{\max}\,(T_{\min}/T_{\max})^r\Big\},
\]
which starts with smooth mixtures for stable optimization and gradually sharpens opacities for discriminative alignment. 
Compared with linear decay, our exponential schedule naturally allocates more iterations at low temperatures (see \cref{fig:temp_decay}), empirically yielding \emph{sharper} per-Gaussian opacities and stronger language alignment while preserving end-to-end differentiability throughout. 
With the proposed temperature-decay schedule, as $\tau$ decreases, mixture blur is gradually reduced and supervision is increasingly transferred to individual Gaussians.

\begin{table*}[t]
    \caption{\textbf{Monocular results on Occ-ScanNet.}
    $^{\dagger}$We re-implement POP-3D~\cite{pop3d} and LOcc~\cite{yu2025language} for the indoor monocular setting, and \textbf{supervise geometry with ground-truth binary occupancy} instead of the authors’ LiDAR-derived pseudo labels. We also adopt DepthAnything-v2~\cite{depthanythingv2} as the backbone and use the same input resolution and training recipe as ours for fair comparison. 
    For POP-3D, we project voxel centers to the image plane and sample language-aligned features at those locations to form a 3D grid of text-aligned embeddings for open-vocabulary reasoning.
    For LOcc, we prompt Qwen2.5-VL-7B~\cite{Qwen2.5-VL} to extract object names and use Trident~\cite{trident} to obtain training-free open-vocabulary segmentations, which yield stronger 2D masks on Occ-ScanNet.
    }
    \vspace{-3mm}
    \setlength{\tabcolsep}{0.008\textwidth}
    \captionsetup{font=scriptsize}
    \begin{center}
    \resizebox{0.96\linewidth}{!}{
    \begin{tabular}{l|c|c c c c c c c c c c c|c}
        \toprule
        Method
        & {IoU}
        & \rotatebox{90}{\parbox{1.5cm}{\textcolor{ceiling}{$\blacksquare$} ceiling}} 
        & \rotatebox{90}{\textcolor{floor}{$\blacksquare$} floor}
        & \rotatebox{90}{\textcolor{wall}{$\blacksquare$} wall} 
        & \rotatebox{90}{\textcolor{window}{$\blacksquare$} window} 
        & \rotatebox{90}{\textcolor{chair}{$\blacksquare$} chair} 
        & \rotatebox{90}{\textcolor{bed}{$\blacksquare$} bed} 
        & \rotatebox{90}{\textcolor{sofa}{$\blacksquare$} sofa} 
        & \rotatebox{90}{\textcolor{table}{$\blacksquare$} table} 
        & \rotatebox{90}{\textcolor{tvs}{$\blacksquare$} tvs} 
        & \rotatebox{90}{\textcolor{furniture}{$\blacksquare$} furniture} 
        & \rotatebox{90}{\textcolor{objects}{$\blacksquare$} objects} 
        & mIoU\\
        \midrule
        \rowcolor{Sbg}\multicolumn{14}{l}{\textit{Closed-vocabulary (full annotation)}}\\
        \arrayrulecolor{lightgray}\midrule\arrayrulecolor{black}
        \rowcolor{Sbg} TPVFormer~\cite{Triformer} & 33.39 & 6.96 & 32.97 & 14.41 & 9.10 & 24.01 & 41.49 & 45.44 & 28.61 & 10.66 & 35.37 & 25.31 & 24.94 \\
        \rowcolor{Sbg} GaussianFormer~\cite{gaussianformer} & 40.91 & 20.70 & 42.00 & 23.40 & 17.40 & 27.0 & 44.30 & 44.80 & 32.70 & 15.30 & 36.70 & 25.00 & 29.93 \\
        \rowcolor{Sbg} MonoScene~\cite{monoscene} & 41.60 & 15.17 & 44.71 & 22.41 & 12.55 & 26.11 & 27.03 & 35.91 & 28.32 & 6.57 & 32.16 & 19.84 & 24.62 \\
        \rowcolor{Sbg} ISO~\cite{ISO} & 42.16 & 19.88 & 41.88 & 22.37 & 16.98 & 29.09 & 42.43 & 42.00 & 29.60 & 10.62 & 36.36 & 24.61 & 28.71 \\
        \rowcolor{Sbg} Surroundocc~\cite{surroundocc} & 42.52 & 18.90 & 49.30 & 24.80 & 18.00 & 26.80 & 42.00 & 44.10 & 32.90 & 18.60 & 36.80 & 26.90 & 30.83 \\
        \rowcolor{Sbg} EmbodiedOcc~\cite{embodiedocc} & 53.55 & 39.60 & 50.40 & 41.40 & 31.70 & 40.90 & 55.00 & 61.40 & 44.00 & 36.10 & 53.90 & 42.20 & 45.15 \\
        \rowcolor{Sbg} EmbodiedOcc++~\cite{embodiedocc++} & 54.90 & 36.40 & 53.10 & 41.80 & 34.40 & 42.90 & 57.30 & 64.10 & 45.20 & 34.80 & 54.20 & 44.10 & 46.20 \\
        \rowcolor{Sbg} RoboOcc~\cite{roboocc} & 56.48 & 45.36 & 53.49 & 44.35 & 34.81 & 43.38 & 56.93 & 63.35 & 46.35 & 36.12 & 55.48 & 44.78 & 47.76 \\
        \midrule
        \rowcolor{Ubg}\multicolumn{14}{l}{\textit{Open-vocabulary (geometry-only annotation)}}\\
        \arrayrulecolor{lightgray}\midrule\arrayrulecolor{black}
        \rowcolor{Ubg} POP-3D$^{\dagger}$~\cite{pop3d} & 35.32 & 0.00 & 23.89 & 16.59 & 0.01 & 4.62 & 8.65 & 2.01 & 7.61 & 0.00 & 2.14 & 0.08 & 5.96  \\
        \rowcolor{Ubg} LOcc$^{\dagger}$~\cite{yu2025language} & 36.70 & 0.00 & 24.40 & 17.37 & 1.83 & 9.31 & 10.73 & 23.42 & 11.29 & 0.01 & 3.15 & 0.19 & 9.25 \\
        \rowcolor{UbgStrong} LegoOcc (Ours) & 59.50 & 17.81 & 41.65 & 28.52 & 18.91 & 19.32 & 33.05 & 34.92 & 19.16 & 5.36 & 5.88 & 6.94 & 21.05 \\
        \bottomrule
    \end{tabular}
    }
    \end{center}
    \vspace{-3mm}
    \label{tab:main_mono}
\end{table*}

\subsection{Losses}\label{sec:loss}
We optimize the network with a composite objective that couples 3D geometry supervision with 2D language-aligned guidance as shown in~\cref{fig:framework}. For occupancy, we follow~\cite{embodiedocc} but use \emph{binary} voxel supervision with focal loss $L_{\mathrm{focal}}$ and Lovász–Softmax $L_{\mathrm{lov}}$, plus a scene-class affinity regularizer $L_{\mathrm{scal}}$ for spatial coherence.
Geometry is further stabilized by a robust Huber depth loss $L_{\mathrm{depth}}$. For semantics, we align image features from LE-Gaussians 
with frozen open-vocabulary segmentation features via a cosine objective $L_{\mathrm{feat}}$. To exploit multi-view consistency without 2D labels, we re-render nearby views (default: 5 frames) and apply the same feature alignment.
The final objective is
\begin{equation}
\label{eq:loss_total}
\begin{aligned}
L_{\mathrm{total}}
\;&=\;
\lambda_{\mathrm{focal}}\,L_{\mathrm{focal}}
\;+\;
\lambda_{\mathrm{lov}}\,L_{\mathrm{lov}}
\;+\;
\lambda_{\mathrm{scal}}\,L_{\mathrm{scal}} \\
\;&+\;
\lambda_{\mathrm{feat}}\,L_{\mathrm{feat}}
\;+\;
\lambda_{\mathrm{depth}}\,L_{\mathrm{depth}}.
\end{aligned}
\end{equation}

\section{Experiments}\label{sec:exp}

\subsection{Datasets and Metrics}

\noindent\textbf{Dataset.}
We train and evaluate on Occ-ScanNet~\cite{ISO}, where each sample provides a $60{\times}60{\times}36$ voxel grid covering a $4.8{\rm m}{\times}4.8{\rm m}{\times}2.88{\rm m}$ volume. The dataset comprises 12 classes in total—11 semantic categories (\emph{ceiling}, \emph{floor}, \emph{wall}, \emph{window}, \emph{chair}, \emph{bed}, \emph{sofa}, \emph{table}, \emph{television}, \emph{furniture}, \emph{objects}) plus \emph{empty}, and is split into 45{,}755 training and 19{,}764 validation samples.

\noindent\textbf{Task setup.}
We study the \emph{monocular} setting, where a single RGB image serves as input.
All experiments on Occ-ScanNet are conducted \emph{without} semantic voxel annotations during training: geometry is supervised by binary occupancy labels, while semantics are learned via language-aligned feature rendering (\cref{sec:gs_temp}).
Evaluation uses the semantic labels for quantitative assessment.

\noindent\textbf{Metrics.}
Following prior work~\cite{ISO,embodiedocc}, evaluation
is restricted to the current camera frustum.
We report: 1) overall occupancy IoU and 2) per-class IoU over the 11 semantic categories, together with the mean IoU (mIoU).

\subsection{Implementation Details}

We adopt Depth-Anything V2~\cite{depthanythingv2} as our depth backbone, which has been used as a geometry prior in~\cite{ISO,embodiedocc}. Building on this, we construct Gaussian primitives using a surface-point expansion strategy similar to~\cite{zhou2026generalizingvisualgeometrypriors}.
For semantics, we use Trident~\cite{trident} as the open-vocabulary segmentation model to extract language-aligned per-pixel features, and the CLIP~\cite{clip} text encoder to embed category names. We follow the standard prompt template ``a photo of $\{\cdot\}$'' to obtain text embeddings.

All models are trained for 10 epochs with a batch size of 4, optimizing with AdamW~\cite{adamw}; the learning rate is initialized at $2\times10^{-4}$ and follows a cosine decay schedule~\cite{coslr} with linear warmup, with weight decay set to $0.01$. We apply gradient clipping~\cite{gradclip} with a maximum norm of $1.0$, and resize each image so that its longer side is $518$ pixels while preserving aspect ratio. We train all our model using 4
NVIDIA GeForce RTX 4090 GPUs.

\subsection{Main Results}\label{sec:main_results}

To assess the effectiveness of our approach, we conduct a comprehensive evaluation on the Occ-ScanNet benchmark~\cite{ISO} and present both overall and per-class Intersection-over-Union (IoU) as well as mean IoU (mIoU) in \cref{tab:main_mono}.
For clarity, rows marked \textcolor[HTML]{2171B5}{blue} denotes \colorbox{Sbg}{\textcolor[HTML]{2171B5}{closed-vocabulary}} (full annotation) settings, while rows in \textcolor[HTML]{C05A00}{orange} denote \colorbox{Ubg}{\textcolor[HTML]{C05A00}{open-vocabulary}} (geometry-only annotation); darker shades highlight our method.

From~\cref{tab:main_mono}, we observe that using geometry-only annotation the re-implemented baselines POP-3D~\cite{pop3d} and LOcc~\cite{yu2025language}
perform poorly, 
obtaining only 35.32/5.96 and 36.70/9.25 IoU/mIoU, respectively. In contrast, our model reaches 59.50 IoU and 21.05 mIoU, outperforming both
by a large margin. Notably, \modelname~achieves the highest IoU among all entries,
indicating that the proposed design effectively recovers occupancy without semantic voxel labels.
The remaining performance gap between full-annotation methods and geometry-only ones in mIoU highlights the intrinsic difficulty of category calibration for open-vocabulary occupancy, particularly under class ambiguity. We further illustrate these problems in our appendix.

\subsection{Ablation Studies}
We use the same coloring convention as in the main results for \colorbox{Sbg}{\textcolor[HTML]{2171B5}{closed-vocabulary}} and \colorbox{Ubg}{\textcolor[HTML]{C05A00}{open-vocabulary}} settings. When a table reports \emph{only} the open-vocabulary setting results, background coloring is omit for brevity. More ablations and analyses are provided in our appendix.

\begin{table}[h]\small
\centering
\vspace{-1mm}
\caption{\textbf{Ablation on the Gaussian-to-Occupancy operator.}
We compare three aggregation rules: GaussianFormer2, Bernoulli, and Poisson, under both \emph{closed-} and \emph{open-} vocabulary settings.}
\vspace{-2mm}
\resizebox{0.6\linewidth}{!}{
\begin{tabular}{c|cc}
\toprule
Operation & mIoU & IoU \\
\midrule
\rowcolor{Sbg} GaussianFormer2 & 51.88 & 56.96 \\
\rowcolor{Sbg} Bernoulli & 51.68 & 58.06 \\
\rowcolor{SbgStrong} Poisson & 52.80 & 59.24 \\
\midrule
\rowcolor{Ubg} GaussianFormer2 & 0.00 & 0.00 \\
\rowcolor{Ubg} Bernoulli & 17.25 & 46.65 \\
\rowcolor{UbgStrong} Poisson & 21.05 & 59.50 \\
\bottomrule
\end{tabular}
}
\label{tab:abs_g2o}
\vspace{-1mm}
\end{table}

\noindent\textbf{Ablation on Gaussian-to-Occupancy (G2O).}
As shown in~\cref{tab:abs_g2o}, replacing the operator from GaussianFormer2~\cite{gaussianformer2} with our Poisson-based G2O consistents improves performance under the \emph{closed-vocabulary} setting, 56.96$\!\rightarrow$59.24 IoU and 51.88$\!\rightarrow$52.80 mIoU.
The advantage is more pronounced in the \emph{open-vocabulary} case, where our Poisson G2O achieves 59.50 IoU and 21.05 mIoU, surpassing the Bernoulli variant by +12.85 IoU and +3.80 mIoU.
These results indicate that incorporating opacity via a Poisson-based compositing framework yields faithful voxel aggregation.

\begin{table}[h]\scriptsize
\centering
\caption{\textbf{Ablation on temperature scheduling.}
We vary the temperature range $(T_{\min}, T_{\max})$, the test-time temperature $\tau_\text{test}$, and the scheduling strategy.
“$\times$” denotes no schedule. %
}
\vspace{-1mm}
\label{tab:abs_temp}
\begin{tabular}{cccc|cc}
\toprule
$T_{min}$ & $T_{max}$ & $\tau_\text{test}$ & scheduler & mIoU & IoU \\
\midrule
1.0 & 1.0 & 1 & $\times$ & 18.15 & 59.19 \\
1.0 & 1.0 & 1e-3 & $\times$ & 12.83 & 32.52 \\
1e-3 & 1e-3 & 1e-3 & $\times$ & 0.00 & 0.00 \\
\midrule
1e-3 & 1.0 & 1e-3 & linear & 2.30 & 7.60 \\
\midrule
1e-2 & 1.0 & 1e-2 & exp & 20.85 & 59.25 \\
1e-3 & 1.0 & 1e-3 & exp & 21.05 & 59.50 \\
1e-4 & 1.0 & 1e-4 & exp & 20.37 & 58.54 \\
\bottomrule
\vspace{-3mm}
\end{tabular}
\end{table}

\noindent\textbf{Ablation on temperature scheduling.}
As reported in \cref{tab:abs_temp}, fixing $\tau{=}1$ throughout training and testing (\ie, no schedule) yields reasonable geometry (59.19 IoU) but poor semantics (18.15 mIoU), indicating substantial per-ray feature mixing degrades semantic discrimination.
Using mismatched temperatures ($\tau_{\text{train}}{=}1$, $\tau_{\text{test}}{=}10^{-3}$)
severely degrades geometry (32.52 IoU) due to a train–test distribution shift. Keeping a constant low temperature ($\tau{=}10^{-3}$) collapses optimization entirely, underscoring the necessity of progressive decay.
A linear decay from $1\!\to\!10^{-3}$ with few iterations near $\tau\!\approx\!0$ also underperforms (2.30 mIoU/7.60 IoU), suggesting that allocating too few steps in the low-temperature regime inadequately sharpen per-Gaussian opacities.
In contrast, our exponential schedule, which allocates more iterations to the low-temperature regime, achieves the best results: 21.05 mIoU/59.50 IoU with $(T_{\min},T_{\max}){=}(10^{-3},1)$ and $\tau_{\text{test}}{=}10^{-3}$.
Lowering $T_{\min}$ to $10^{-4}$ slightly hurts performance (20.37 mIoU/58.54 IoU), suggesting over-sharpening amplifies noise. 
These results demonstrate that progressive temperature decay effectively reduces feature mixing and yields sharper, more discriminative Gaussian embeddings.

\begin{table}[h]\small
\centering
\caption{\textbf{Model profile.} FPS is measured on the same machine with a single RTX 4090 GPU, averaged over 1{,}000 runs after 100 warm-up runs.}
\vspace{-1mm}
\resizebox{0.96\linewidth}{!}{
    \begin{tabular}{lcccc}
    \toprule
    \textbf{Model} & mIoU & IoU & FPS & \#parameters \\
    \midrule
    \rowcolor{Sbg} ISO~\cite{ISO} & 28.71 & 42.16 & 3.81 & 303.05M \\
    \rowcolor{Sbg} EmbodiedOcc~\cite{embodiedocc} & 45.15 & 53.55 & 11.48 & 231.45M \\
    \midrule
    \rowcolor{Ubg} POP-3D~\cite{pop3d} & 5.96 & 35.32 & 10.21 & 115.31M \\
    \rowcolor{Ubg} LOcc~\cite{yu2025language} & 9.25 & 36.70 & 8.93 & 113.50M \\
    \rowcolor{UbgStrong} Ours & 21.05 & 59.50 & 22.47 & 99.04M \\
    \bottomrule
    \end{tabular}
}
\label{tab:fps}
\vspace{-1mm}
\end{table}

\noindent\textbf{Model profile.} We summarize the model profile in~\cref{tab:fps}. Note that ISO~\cite{ISO} and EmbodiedOcc~\cite{embodiedocc}
rely on closed-set classification heads, while open-vocabulary methods compare occupancy features with language embeddings~\cite{clip} via cosine similarity.

\begin{figure*}[h]
  \centering
  \includegraphics[width=0.98\textwidth]{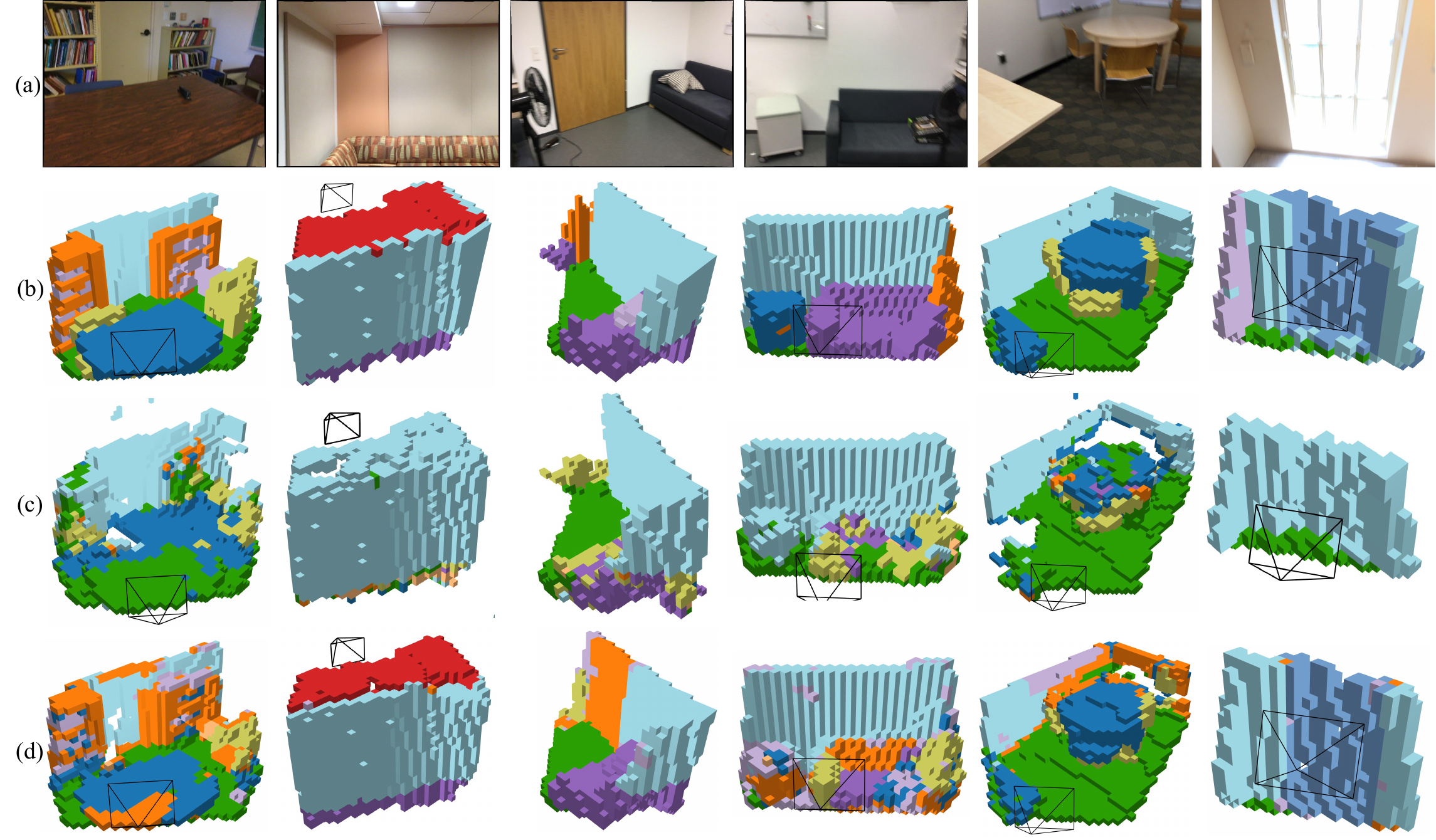}
  \vspace{-1mm}
  \caption{\textbf{Qualitative results on Occ-ScanNet.} From top to bottom: (a) input images; (b) ground-truth semantic occupancy; (c) results from our re-implemented LOcc~\cite{yu2025language}; (d) our method. Both (c) and (d) are trained with geometry-only annotations and evaluated on the closed-vocabulary annotation of Occ-ScanNet.}
  \vspace{-1mm}
  \label{fig:quan_sl}
\end{figure*}

\begin{figure*}[h]
  \centering
  \includegraphics[width=0.98\textwidth]{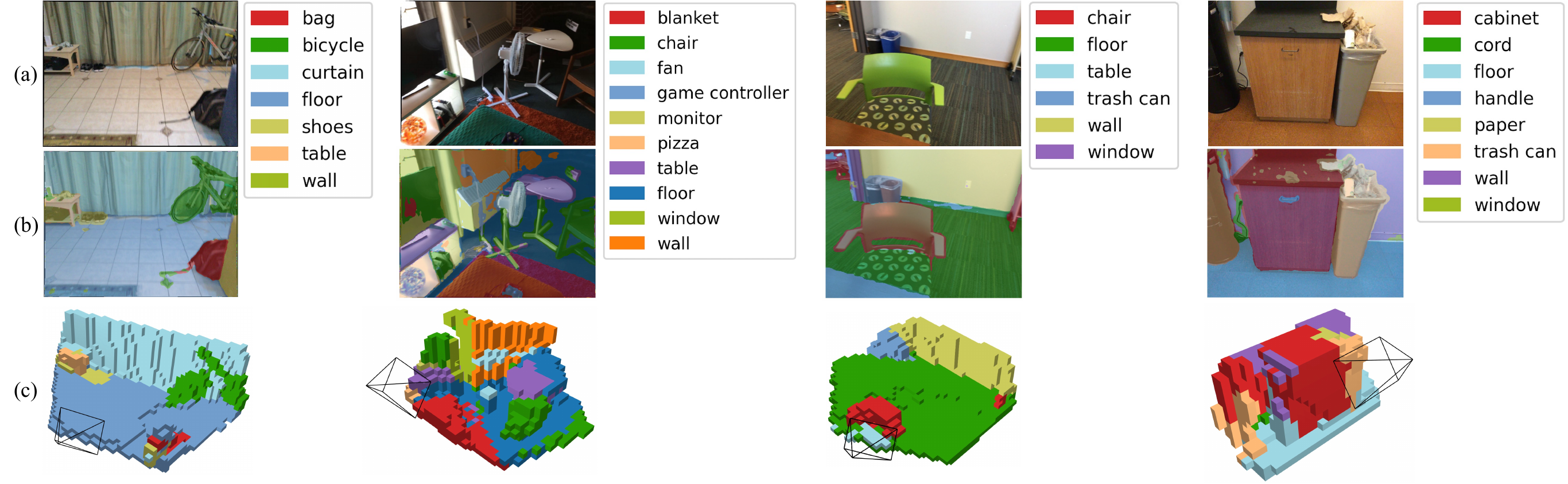}
  \vspace{-1mm}
  \caption{\textbf{Open-vocabulary qualitative results.}
  Legends list the \emph{VLM-extracted object nouns} used as text queries.
  (a) Input image. (b) Open-vocabulary 2D segmentation for queried nouns. 
  (c) Our 3D open-vocabulary occupancy colored by the same categories.}
  \vspace{-1mm}
  \label{fig:quan_ov}
\end{figure*}

\subsection{Qualitative Visualization}

We present qualitative results under both closed and open vocabulary settings. As shown in~\cref{fig:quan_sl}, our open-vocabulary model is evaluated against the closed-set semantics of Occ-ScanNet~\cite{ISO}. Across diverse scenes, the predicted occupancy aligns closely with ground-truth labels, indicating strong 
geometric accuracy and semantic generalization to the dataset taxonomy.
Beyond this closed set, \cref{fig:quan_ov} presents open-vocabulary predictions obtained by
querying arbitrary given categories. For each image, a VLM~\cite{Qwen2.5-VL} extracts object \emph{nouns} as text queries, which our model uses to produce the corresponding 3D semantic occupancies.
More qualitative visualizations are provided in our appendix, including text-conditioned 3D grounding examples.
These results demonstrate the model's ability to localize and identify free-form indoor categories in 3D.

\section{Conclusion}
We introduced a monocular open-vocabulary occupancy framework for large-scale indoor scenes, where category distributions are long-tailed and open-ended. Central to our approach is a reformulated, opacity-aware Poisson-based Gaussian-to-Occupancy operator that enables Gaussians to serve as a unified semantic–geometric intermediate representation. Complementing this, we propose a Progressive Temperature Decay schedule to mitigate feature dilution during splatting and enhance discriminability. Extensive experiments on Occ-ScanNet demonstrate strong performance. Our method supports text-conditioned queries and scales across embodied tasks without closed-vocabulary constraints, paving the way for wider occupancy use in real-world embodied applications.

\section*{Acknowledgements}

This work was supported by National Natural Science Foundation of China (NFSC) under the Grant Number 62573370 and Key Area Project of Education Department of Guangdong Province (No. 2025ZDZX3051).

{
    \small
    \bibliographystyle{ieeenat_fullname}
    \bibliography{main}
}

\end{document}